\documentclass[letterpaper, 10 pt, conference]{ieeeconf}  

\IEEEoverridecommandlockouts                              

\usepackage{xcolor}
\usepackage{graphicx}
\usepackage{subcaption}

\title{\LARGE \bf
A HeARTfelt Robot: Social Robot-Driven Deep Emotional Art Reflection with Children

}

\author{Isabella Pu$^{1}$, Golda Nguyen$^{2}$, Lama Alsultan$^{1,3}$, Rosalind Picard$^{1}$, Cynthia Breazeal$^{1}$, Sharifa Alghowinem$^{1}$
\thanks{$^{1}$ MIT Media Lab,
        Massachusetts Institute of Technology, Cambridge, MA, United States,
        {\tt\small ipu@media.mit.edu}}%
\thanks{$^{2}$ Department of Aeronautics and Astronautics,
        Massachusetts Institute of Technology, Cambridge, MA, United States}%
\thanks{$^{3}$ Department of Computing and Information Systems,
        Prince Sultan University, Riyadh, Saudi Arabia}%
}

\begin{document}

\maketitle
\thispagestyle{empty}
\pagestyle{empty}



\begin{abstract}

Social-emotional learning (SEL) skills are essential for children to develop to provide a foundation for future relational and academic success. Using art as a medium for creation or as a topic to provoke conversation is a well-known method of SEL learning. Similarly, social robots have been used to teach SEL competencies like empathy, but the combination of art and social robotics has been minimally explored. In this paper, we present a novel child-robot interaction designed to foster empathy and promote SEL competencies via a conversation about art scaffolded by a social robot. Participants (N=11, age range: 7-11) conversed with a social robot about emotional and neutral art. Analysis of video and speech data demonstrated that this interaction design successfully engaged children in the practice of SEL skills, like emotion recognition and self-awareness, and greater rates of empathetic reasoning were observed when children engaged with the robot about emotional art. This study demonstrated that art-based reflection with a social robot, particularly on emotional art, can foster empathy in children, and interactions with a social robot help alleviate discomfort when sharing deep or vulnerable emotions. 

\end{abstract}
\section{INTRODUCTION \& BACKGROUND}
Social and emotional intelligence is an essential skill for individuals to effectively communicate, interact, and build relationships. This can be fostered through \textbf{social-emotional learning}, which Elias et al. describes as the systematic acquisition of emotional intelligence by developing relevant skills, attitudes, and values \cite{elias1997promoting}. The Collaborative for Academic, Social, and Emotional Learning (CASEL) names core SEL skills as self-awareness, self-management, social awareness, relationship skills, and responsible decision-making \cite{skoog2020evidence}. 

Early exposure to SEL provides significant short-term and long-term behavioral benefits, such as improved self-confidence \cite{greenberg2017social}, reduced likelihood of emotional issues and conduct problems \cite{mondi_fostering_2021, hawkins2008effects}, and improved long-term academic and relational success \cite{taylor_promoting_2017, durlak_impact_2011}. Additionally, a 2013 survey showed 97\% of teachers acknowledge the positive impact of SEL on students from all socioeconomic backgrounds \cite{bridgeland2013missing}, and a study from the Aspen Institute found that SEL programs were particularly beneficial for fostering well-being in children from low-income communities \cite{aspen2019nation}. 

\subsection{Social Robots for Social-Emotional Learning}
To increase access to SEL programming, interactive technologies can deliver educational content inside and outside of the classroom. Social robots and artificial intelligence systems, through responsive engagement with interaction partners, are becoming increasingly prevalent in early childhood education \cite{kanda2012children, thomaz2013active}. However, social robots have predominantly been used in childhood education for academic learning, focusing on areas like language \cite{kanero2018social, van2019social}, literacy, \cite{neumann2020social, chen2020teaching}, and computer science skills \cite{williams2018popbots, rafique2020computation}, while social robots for SEL practice have been relatively limited.

Social robots are a promising method of delivering early SEL education to children given children's high engagement with these technologies \cite{westlund2017children, fridin2014storytelling}. Previous work has shown social robots can alleviate anxiety in children by offering reassurance, and that children felt comfortable sharing emotions with a social robot \cite{dosso2023safe}. Another promising capability of social robots is their ability to convey empathy and to foster the development of empathy in children \cite{pashevich2022can, spitale2022socially}.

Though fewer in number compared to studies on academic education with social robots, prior work has demonstrated success in using social robots to teach SEL competencies. 
Embodied's Moxie, a social robot designed to teach SEL skills to children with developmental disorders, significantly improved children's SEL compentencies \cite{hurst2020social}. 
Several prior studies have focused on teaching emotion recognition and empathy to neurodivergent children (i.e., with autism spectrum disorder) \cite{kewalramani2023scoping,marino2020outcomes,yun2017social,wolfe2018deploying}, but SEL training is impactful for both neurodivergent and neurotypical populations. 







\subsection{Social-Emotional Learning with Art}
Using art as a medium for self-expression or to evoke thoughtful conversation is a widely recognized method of SEL instruction \cite{elias1997promoting, eisner2002arts, brouillette2009arts}. Prior studies have demonstrated the use of both art education \cite{farrington2019arts} and artistic creation \cite{cooney2018design, cooney2021robot} to build interpersonal and social skills, like self-awareness \cite{skoog2020evidence, eisner2002arts}, that contribute to overall emotional development in children \cite{heath1998living}. 


Different mediums of art have also been explored for SEL programming, including dance, music, and visual arts \cite{eddy2021local}. While most arts curricula for teaching SEL involve creating art, several examples have focused on discussing and observing art. Ebert et al. \cite{ebert2015teaching} conducted a workshop where children observed emotions in subjects of different artworks to develop SEL skills like emotion recognition and self-awareness. The Metropolitan Museum of Art \cite{met_curriculum} developed a curriculum based on art observation and discussion to help students practice self-awareness, self-management, social awareness, empathy, and relationship skills---this curriculum was specifically designed for blind or partially sighted students, students with autism spectrum disorder (ASD), and students with developmental disabilities. In these curricula, children successfully developed skills like empathy and emotion recognition by discussing art in scaffolded settings.


\subsection{Study Objectives}
The use of art has been minimally explored in tandem with interactive and responsive social agents for SEL instruction. Cooney et al. \cite{cooney2018design, cooney2021robot} demonstrated the potential of social robots for art therapy to facilitate emotional expression via artistic creation. We expand on this connection by exploring robot-guided conversations on art, in which art is used as a conversational catalyst for children to practice emotion recognition, self-awareness, and empathy. Specifically, an experimental study was conducted to explore how different styles of art (explicitly emotive art and neutral art) affect behavioral responses in children (ages 7-11). Reflection on the artwork was scaffolded by guiding questions and responsive dialogue from a social robot, Jibo \cite{jibo}, to investigate the following research questions: 
\begin{itemize}
    \item \textbf{RQ1: }When interactively reflecting with a social robot, does emotional art foster empathy more successfully than neutral art in children? %
    \item \textbf{RQ2: }Are there behavioral differences (in engagement, disclosure) when children reflect with a social robot on emotional art versus neutral art?
\end{itemize}

\noindent We propose the following hypotheses:
\begin{itemize}
    \item \textbf{H1: }Emotional art will foster more empathy than neutral art during interactive reflection with a social robot.
    \item \textbf{H2: }Children will engage more deeply with emotional art than neutral art during the social robot interaction.
    \item \textbf{H3: }Children who are more open when sharing feelings with the social robot will engage more with the activity.
\end{itemize}

\section{INTERACTION DESIGN}

\subsection{Robot Station Design}
  \begin{figure}[thpb]
      \centering
      \includegraphics[scale=0.325]{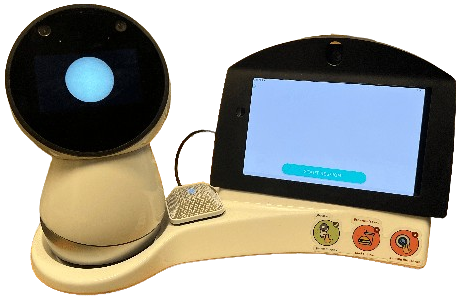}
      \caption{Jibo Robot Station}
      \label{fig:station}
   \end{figure}

 \begin{figure}[thpb]
      \centering
      \includegraphics[scale=0.0655]{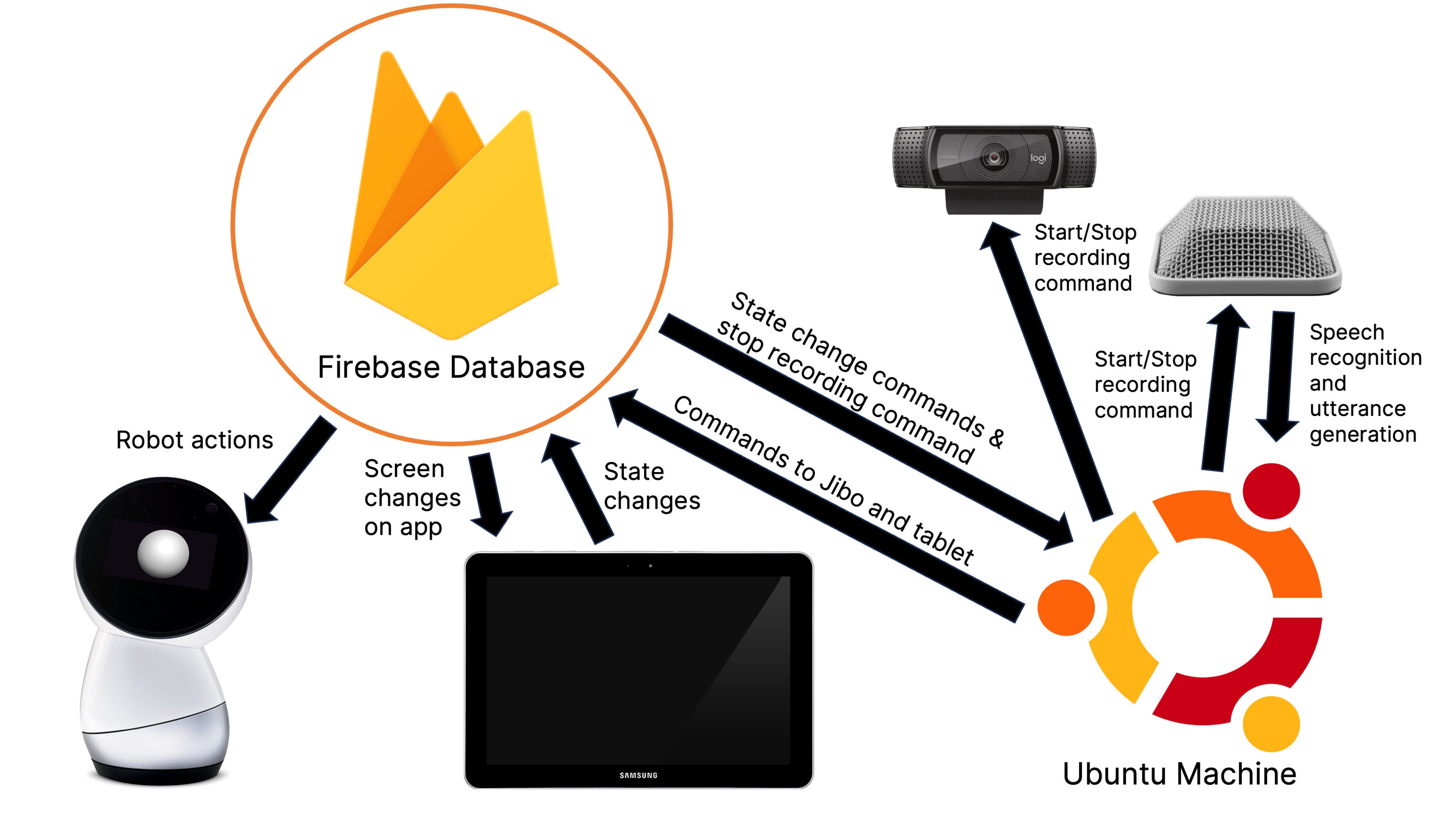}
      \caption{Robot Station System Architecture}
      \label{fig:arch}
   \end{figure}
   
We designed an interaction where participants spoke with Jibo \cite{jibo}, a social embodied agent, through a multi-device system (the ``Robot Station'', shown in Figure \ref{fig:station}). Jibo was used because of its ability to express emotional states with bodily animations, expressive screen-based face, child-friendly character design, and durability for autonomous conversational interactions. The Robot Station has a Jibo robot on the left side and a Samsung Galaxy tablet on the right. There is also a Logitech C930e camera housed above the tablet and a separate MXL AC-44 microphone located between Jibo and the tablet. An Ubuntu machine located inside the Robot Station behind the tablet controls Jibo's movements and utterances.

The Robot Station design allows for Jibo to look between the participant and the tablet for social engagement. Jibo can also freely rotate around three axes while placed in the Robot Station, to provide emotive movement while speaking. 


Jibo converses with the participant by asking scripted questions about artwork and replying with generations from GPT-4\footnote{https://openai.com/research/gpt-4}, in response to participant dialogue. The Ubuntu machine sends commands to Jibo through messaging a Firebase database, which Jibo reads from. The tablet also communicates with the same database, allowing Jibo, the Ubuntu machine, and the tablet to execute synchronously. The system architecture is shown in detail in Figure \ref{fig:arch}. 


\subsection{Art Design for Robot Interaction}
Two categories of artwork were used to prompt empathy and disclosure in the child-robot interaction:  1) \textit{emotional} and 2) \textit{neutral} art. Artworks shown to participants were generated with DALL-E 3\footnote{https://openai.com/dall-e-3} and reviewed in advance by the research team to ensure their suitability for children. 

   \begin{figure}[thpb]
      \centering
      \includegraphics[scale=0.068]{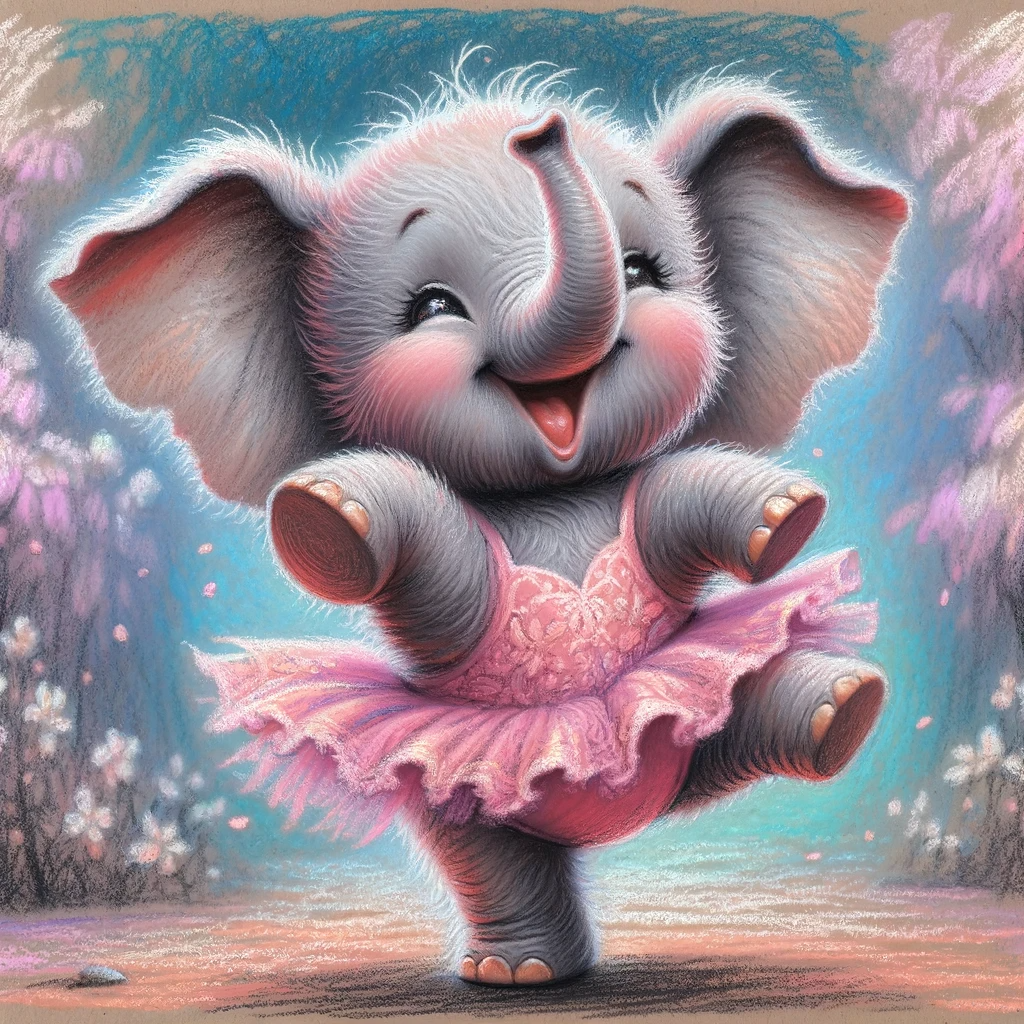}
      \includegraphics[scale=0.068]{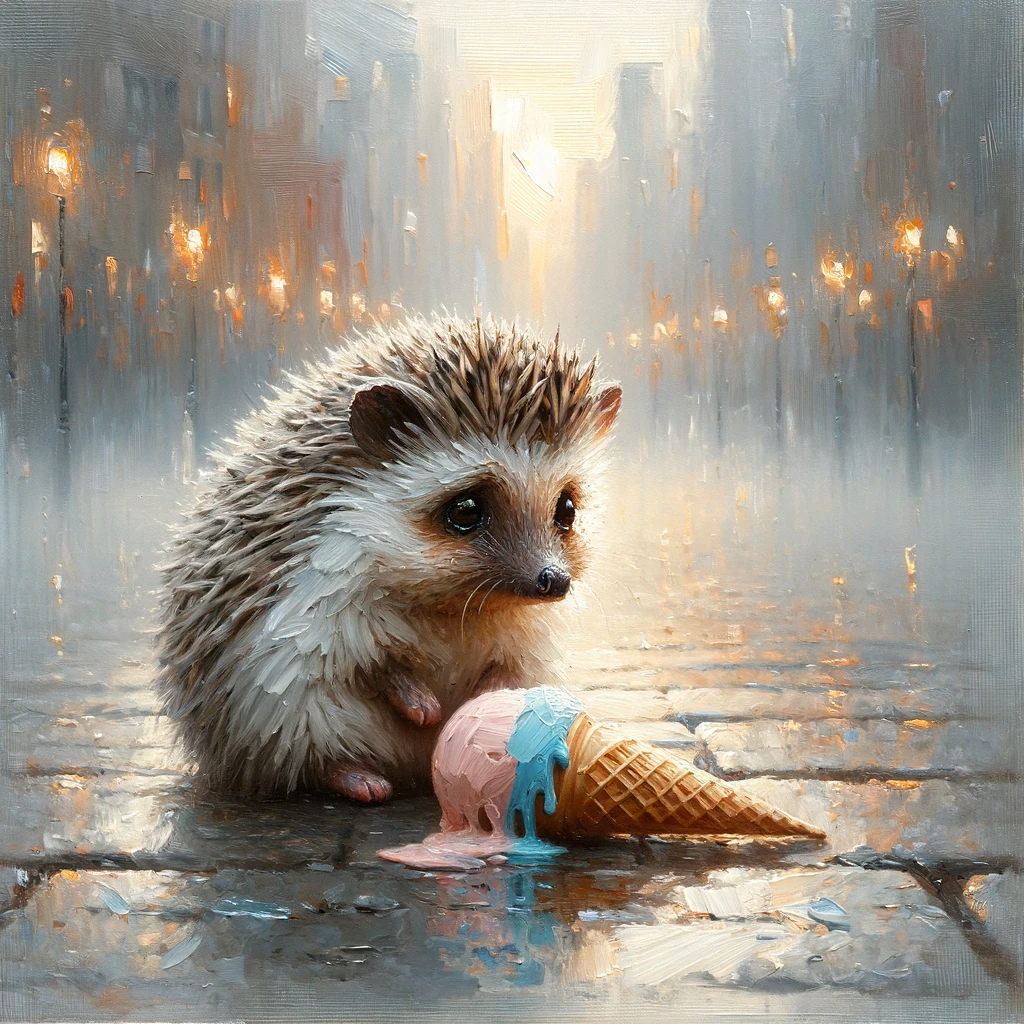}
      \includegraphics[scale=0.068]{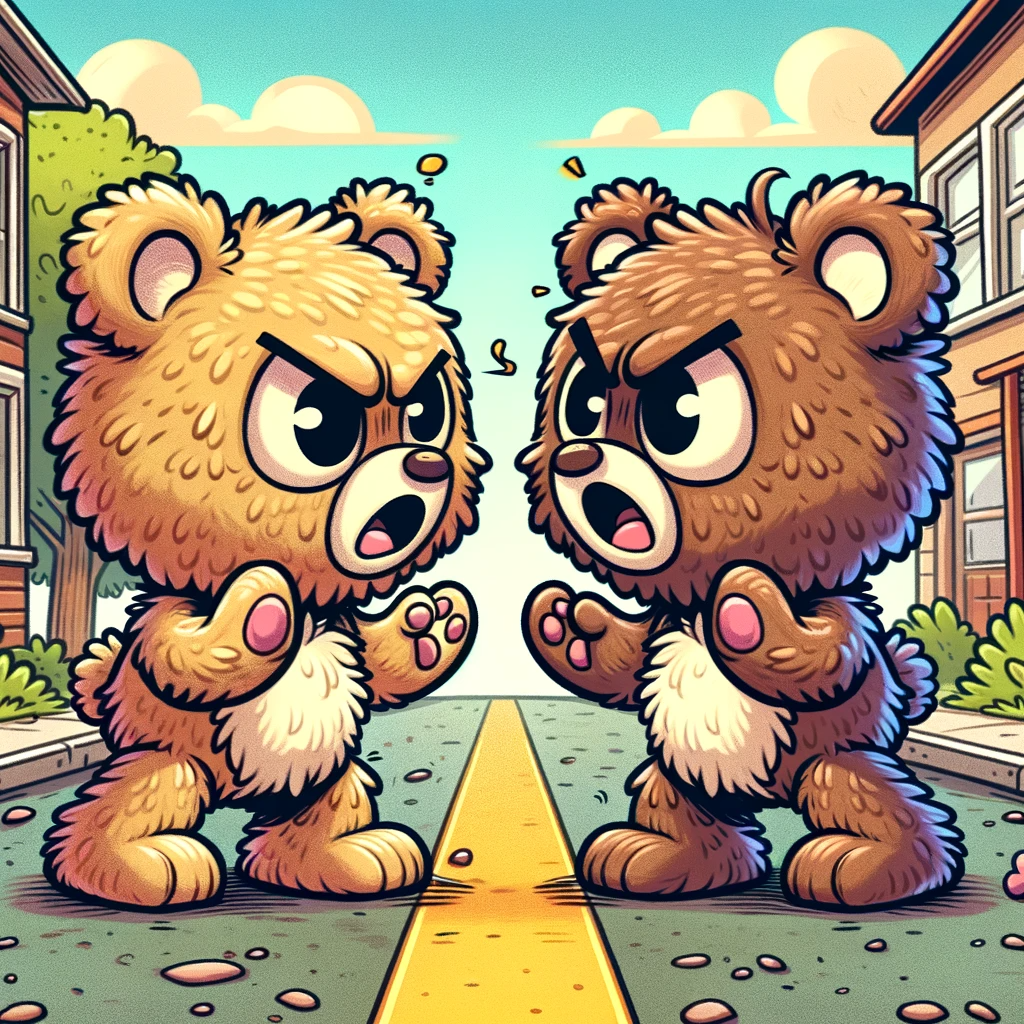}
      \caption{Emotional artworks}
      \label{fig:emo_art}
   \end{figure}


  \begin{figure}[thpb]
      \centering
      \includegraphics[scale=0.068]{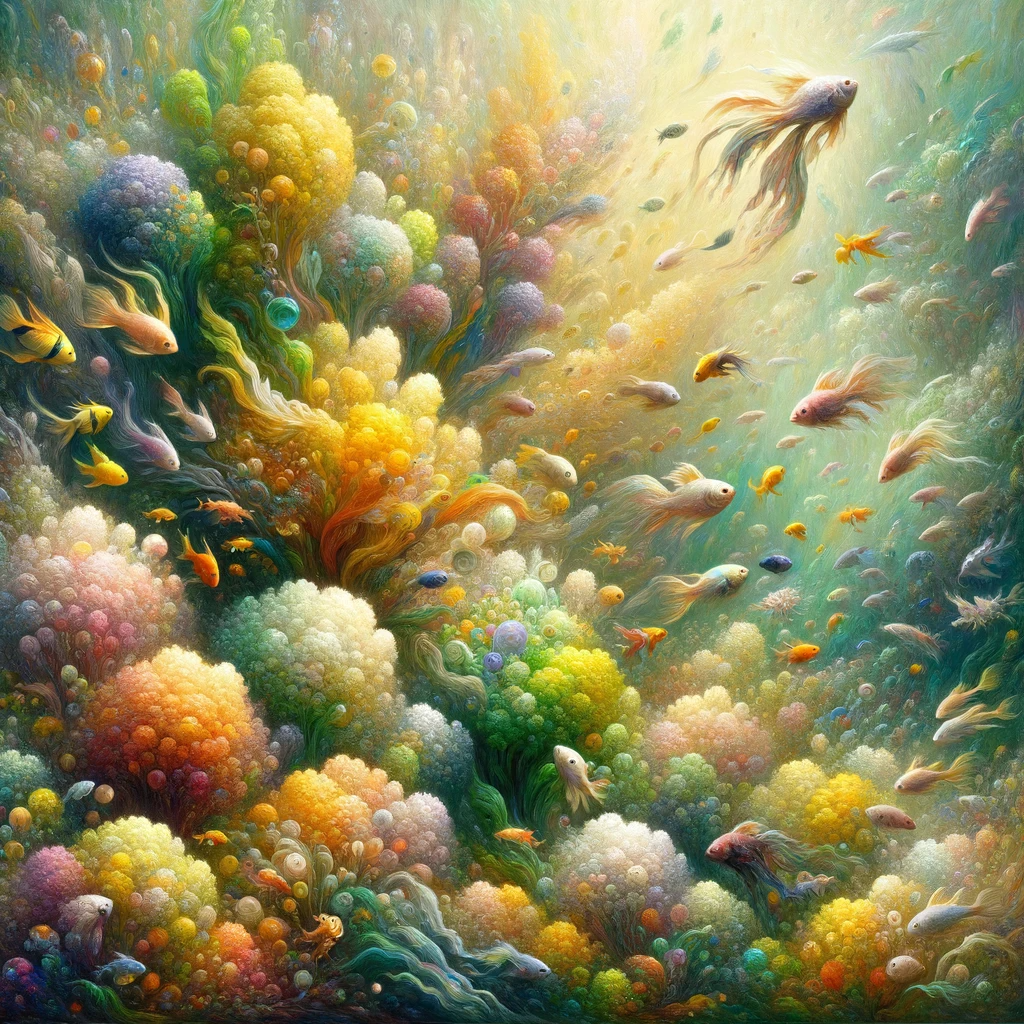}
      \includegraphics[scale=0.068]{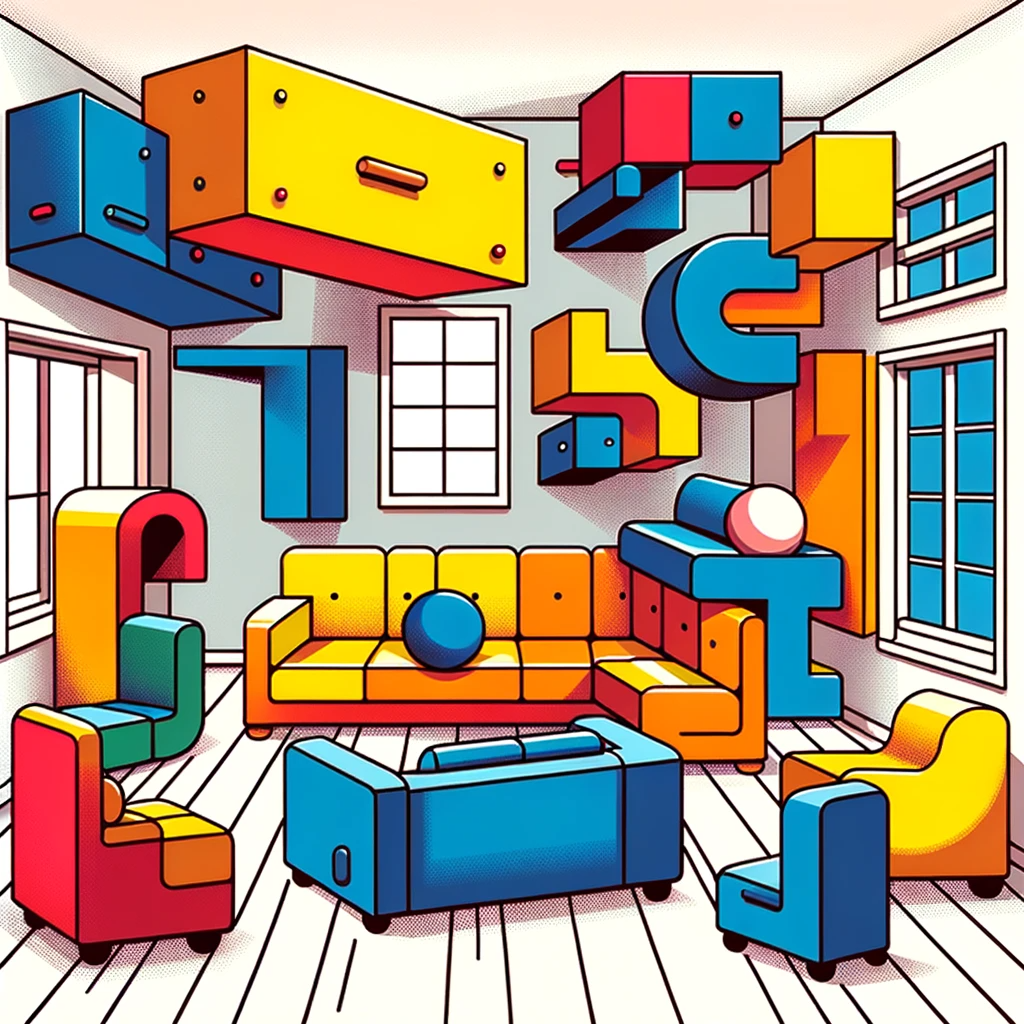}
      \includegraphics[scale=0.068]{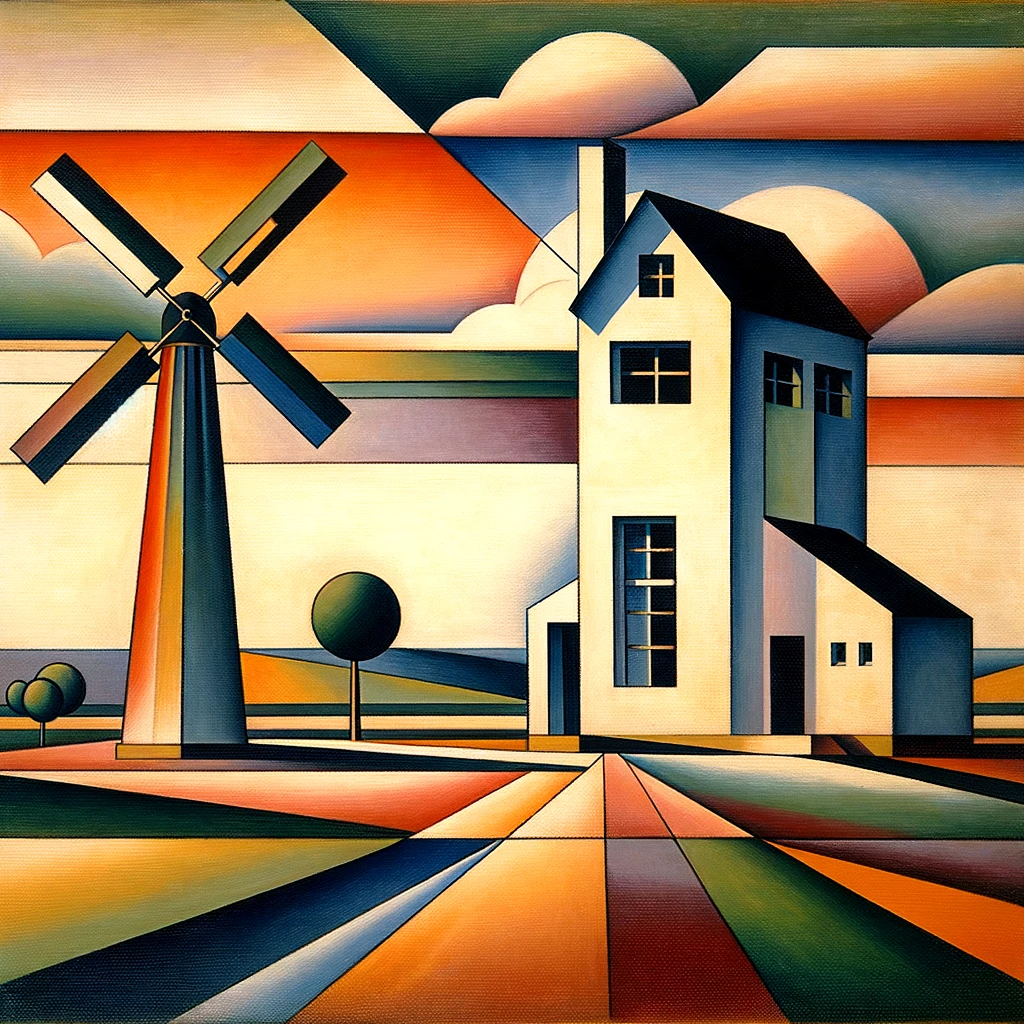}
      \caption{Neutral artworks}
      \label{fig:neutral_art}
   \end{figure}


\textit{Emotional} artworks (Figure \ref{fig:emo_art}) featured animal characters explicitly displaying an emotion: happiness, sadness, or anger. Different styles were used for variety but all included vibrant colors and animals to appeal to children. As an example, the prompt for the ``anger'' image was: \textit{Two angry teddy bears arguing in the street, kid-friendly comic style, hand-drawn, with vibrant colors}.

\textit{Neutral} artworks (Figure \ref{fig:neutral_art}) depicted abstract or landscape imagery to elicit emotions, but not any specific emotion, based on color, composition, or style. These pieces were also designed for children, with bright colors and familiar locations (the ocean, a living room, and a farm). For example, the ``living room'' image prompt was: \textit{Memphis Group style image of a colorful living room where the furniture is upside down, sideways, and backwards, with a blue couch that is right side up. 2D style with bold shapes}.

We validated the emotional effects of these generated artworks during the experimental study (see Section IV).

\section{EXPERIMENT \& ANALYSIS METHODOLOGY}
An experimental study was conducted to explore how social robots can help children \textbf{foster empathy} and \textbf{practice SEL skills} via conversation on art. The study consisted of two sessions: 1) the participant discussed ``neutral art'' (without clear emotions) with a social robot (Jibo), and 2) the participant discussed ``emotional art'' (where characters in the art explicitly convey emotions) with the social robot. A within-subjects design was used, and session order was randomized to control for ordering effects. The study protocol was approved by our institution's ethics review board. 

\subsection{Experimental Procedure}
Participants and their parents were invited to an in-person study in an enclosed space. Two members of the research team were also in the room. The study lasted approximately one hour with two robot interaction sessions (each about 15 minutes), with a break between sessions. Sessions were video- and audio-recorded using two cameras (the robot station and a wall-mounted GoPro) and the station microphone.

The first interaction session with Jibo began with a brief tutorial on how to record responses to Jibo's questions and two neutral practice questions, asking for the participant's name and age. After the tutorial, participants engaged in the first interaction session (one of the two categories described in \textit{Art Design for Robot Interaction}). There was a short break after the first session before participants began the second session. After both sessions were completed, participants were interviewed on their experience.

Within each session, Jibo asked the following questions about each image: 
\begin{enumerate}
    \item Can you tell me a story about this picture or describe this picture to me?
    \item What emotion does this picture make you feel?
    \item Why does this picture make you feel that emotion?
    \item Can you tell me about the last time you felt that emotion?
\end{enumerate}

The questions were designed to guide participants to reflect on emotions they observed or related to within the art piece. Specifically, the questions targeted self-awareness (``identifying one's emotions'', ``linking feelings, values, and thoughts'') and relationship skills (``communicating effectively'') in all sessions, while additionally targeting social awareness (``demonstrating empathy and compassion'') in the emotional session \cite{skoog2020evidence}. Table \ref{tab:conv_ex} provides an example of question prompts from Jibo and P-10's responses to an emotional image. 

\begin{table}[htb]
\caption{Example Interaction between Jibo and P-10}
\label{tab:conv_ex}
\begin{tabular}{|p{0.325\linewidth}|p{0.575\linewidth}|}
\hline
\multicolumn{1}{|c|}{\textbf{Jibo}:} & \multicolumn{1}{c|}{\textbf{P-10}:} \\ \hline
Can you tell me a story about this picture or describe this picture to me? & The picture is about two bears that probably got in a fight. And they wanted... and they’re like, mad at each other. \\ \hline
What emotion does this picture make you feel? & It makes me feel sort of... It's sort of silly. It makes me feel sort of silly, and scared. It makes me feel scared because I don't like when people get in fights, and it makes me feel sort of silly because it's funny that there are two bears that are in a fight. \\ \hline
Why does this picture make you feel that emotion? & Well, it's because I don't like when people get hurt and fighting usually means that you can get hurt. But it's funny to watch. Because, I don't know. I just feel this emotion. \\ \hline
Can you tell me about the last time you felt that emotion? & Well, one time I was with my brother and I got in a fight with him because I got mad at him. We didn't get hurt, but I got scared. And I felt like it was sort of silly that we were both fighting though. \\ \hline
\end{tabular}
\end{table}

Jibo's utterances were generated by GPT-4 to adapt to participant responses for a more personalized interaction (GPT-4 responses are not shown in Table \ref{tab:conv_ex} for brevity). 
For example, a participant told Jibo about a new pet gecko, and Jibo's response referenced the gecko and gave an encouraging comment on the bond between humans and pets.

Though GPT-4 responses may vary between participants due to the AI's nature, using the same strict and detailed prompts for all participants ensured that the responses maintained consistent vocabulary and were within the same vein.

The post-study interview was conducted without the Jibo robot present, to minimize potential effects of the robot's presence effect on the child's opinions. The child was asked questions on a tablet about what they liked about the interactions with Jibo, what they disliked, if they would change anything, their feelings of comfort with or trust in Jibo, and if they would want to interact with Jibo again.

\subsection{Participants}
This study was conducted with 11 children between age 7 and 11 (average age of 9.3), with 6 female and 5 male participants. 7 participants were white, 2 were Asian, and 2 were mixed race. Ethnicity was not considered as a factor in this study and was not analyzed any further than for demographic purposes. Legal guardians were consented, and participants provided their assent.

Participants were recruited via email advertisements to parents of children who had previously participated in community outreach programs or indicated interest in robot and AI studies. Participants and their family were not familiar with the specific researchers facilitating this study. Participants' travel to the study location was reimbursed, but they were not otherwise compensated \cite{cordero2022review}.

\subsection{Linguistic Analysis Methods}
Linguistic analysis of participant dialogue was performed to collect measures of conversational engagement and disclosure. Speech was first transcribed by Assembly AI Automatic Speech Recognition\footnote{https://www.assemblyai.com/}, then manually cleaned by the research team. Common themes were coded by question using thematic content analysis to examine emotional trends. Coding was performed by two independent raters, and inter-rater agreement was calculated using Cohen's Kappa ($\kappa$) \cite{mchugh_interrater_2012}.


Raters examined responses to question \(2)\) and listed all emotions participants cited in order to validate the success of the emotional artwork in eliciting the target emotion.

 Responses to question \(3)\) were coded as exhibiting ``empathetic reasoning'', ``visual reasoning'', or neither.
 Examples of empathetic reasoning included \textit{``... she's happy—just like I am—because she's probably happy that she's dancing because I'm happy when I dance.''} from P-10 and \textit{``It makes me feel upset because the hedgehog has dropped its ice cream, and it looks like really good ice cream.''} from P-06.
Examples of visual reasoning from the same participants included \textit{``It's the theme of the artwork, and also the colors because it has rainbow colors on it and I like rainbow colors.''} from P-10 and \textit{``Because of the way it looks.''} from P-06.
Responses like \textit{``I don't know''} or incoherent responses were labeled as exhibiting neither type of reasoning. This coding had very strong inter-rater agreement ($\kappa=0.82$).
 

 

\subsection{Video Analysis Methods}
To conduct behavioral analysis, video data of the sessions was annotated by two independent annotators using ELAN 6.7, with a coding manual on three measurements: Comfort (ease, relaxation, and lack of anxiety), Engagement (attention, interest, and active participation), and Openness (open, honest, and vulnerable behavior). Each of the measurements included a negative (-1), neutral (0), or positive (1) rating. Annotators specifically marked events with negative (-1) or positive (1) behavior, leaving unmarked sections of video scored as neutral (0). Cohen’s Kappa was calculated on 30\% of the video data to measure inter-rater reliability, revealing a substantial agreement score $(0.61 < \kappa < 0.80)$ across all three measures. Annotators also noted behavioral patterns that stood out or occurred often among participants.


\section{VALIDATION OF EMOTIONAL ARTWORK}

The emotional art used in this study was generated specifically for this experiment, with the goal of having participants empathize with that emotion. The three images used, as seen in Figure \ref{fig:emo_art}, sought to demonstrate happiness, sadness, and anger respectively (viewed from left to right).

Most participants (8 out of 11) referenced some dimension of happiness (synonyms included ``joyful'' and ``excited'') when viewing the image designed to elicit happiness. The other three participants referenced feeling ``fine'' (later elaborating that the image did not elicit a strong feeling), ``weirded out'' (describing that \textit{``a ballerina dress on an elephant seems weird''}), and ``fear'' (explaining that it reminded them of a dancing-related memory that induced fear when recollected).

All participants cited some dimension of sadness (synonyms included ``feeling bad'' and ``upset'') when viewing the image designed to induce sadness. 

The picture designed to induce anger was the most polarizing, with 7 out of 11 participants describing it as making them feel either ``angry'' or ``annoyed''. Of the other 4 participants, 3 expressed that they found the image ``funny'' or that it made them feel ``silly''. When elaborating on their reasoning, these participants explained that they found watching others fight to be a funny or silly experience. However, 2 of these 3 participants still shared a related memory that involved a fight and described having negative emotions during that memory, like fear and confusion. One also mentioned that the image did not elicit anger but reminded them of anger, since the bears depicted in it were fighting. The last participant who did not mention anger instead expressed that they felt ``tired'', likely due to it being the final image they were viewing over the entire study.

Overall, a majority of participants experienced the intended emotion of each picture in the emotional session.

In the neutral session, participants cited more varied emotions (8, 10, and 10 total emotions cited, respectively, for each neutral image). The most common feelings cited in the neutral session were ``creative'', ``energetic'', and ``curious''.

\section{RESULTS}

\subsection{Emotional Art Leads to Empathy and Vulnerability}

\begin{figure*}[htb]
  \centering 
  \begin{subfigure}{.5\textwidth} 
    \centering 
    \includegraphics[width=.95\linewidth]{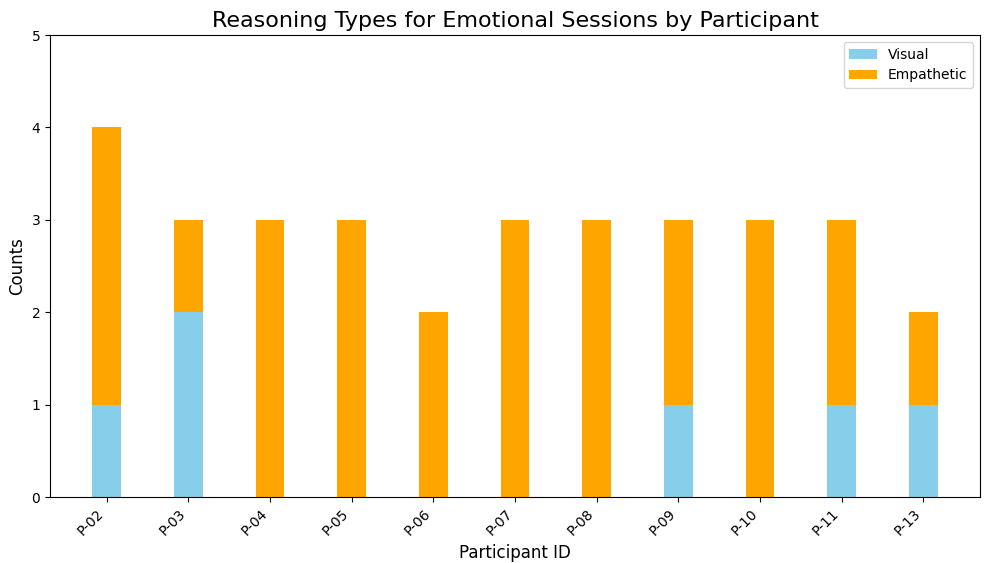} 
    \caption{Emotional sessions}
    \label{fig:e_reasonings}
  \end{subfigure}%
  \begin{subfigure}{.5\textwidth}
    \centering 
    \includegraphics[width=.95\linewidth]{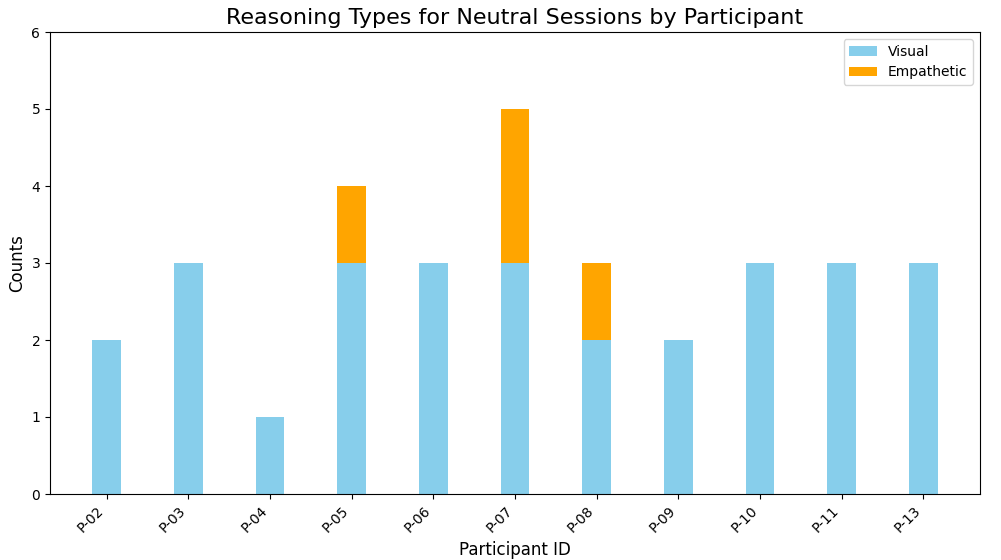} 
    \caption{Neutral sessions}
    \label{fig:n_reasonings}
  \end{subfigure}
  \caption{Empathetic and visual reasoning between sessions}
  \label{fig:reasonings}
\end{figure*}

Participant responses to question \(3)\) on why they felt a certain emotion after viewing an image were coded as containing either ``empathetic reasoning", ``visual reasoning", both, or neither. To investigate \textbf{H1}, counts were performed of how many instances each reasoning type (empathetic reasoning or visual reasoning) was used by each participant per session, and these counts are shown in Figure \ref{fig:reasonings}.

A distinct pattern quickly emerged, showing participants exhibited more empathetic reasoning when viewing emotional art, as opposed to exhibiting more visual reasoning when viewing neutral art. On average in the emotional sessions, participants used empathetic reasoning in 2.4 out of 3 images, while only using visual reasoning in 0.5 out of 3 images. On average in the neutral sessions, participants used empathetic reasoning in 0.4 out of 3 images while using visual reasoning in 2.5 out of 3 images. 


To compare between sessions, Wilcoxon signed-rank tests (non-parametric tests appropriate for small sample sizes) were performed. Significant differences were found, showing significantly more coded instances of empathetic reasoning in the emotional session and significantly more coded instances of visual reasoning in the neutral sessions ($p<.001$), strongly supporting \textbf{H1}.

Additionally, empathy felt by participants during the activity appeared to remain even past the end of the study. For example, during the post-activity interview, P-02 stated that their least favorite part of the activity was \textit{``seeing the sad hedgehog with the dropped ice cream''}.

Participants also appreciated the chance to talk about their emotions, with P-13 stating \textit{``My favorite part of the activities today were the questions that most people wouldn't have asked me, about your feelings...''} and P-10 responding that \textit{``Probably talking with Jibo and saying my emotions about the different artwork...''} was their favorite part of the activity. 

\subsection{Participant Verbosity}

    \begin{figure}[thpb]
      \centering
      \includegraphics[scale=0.35]{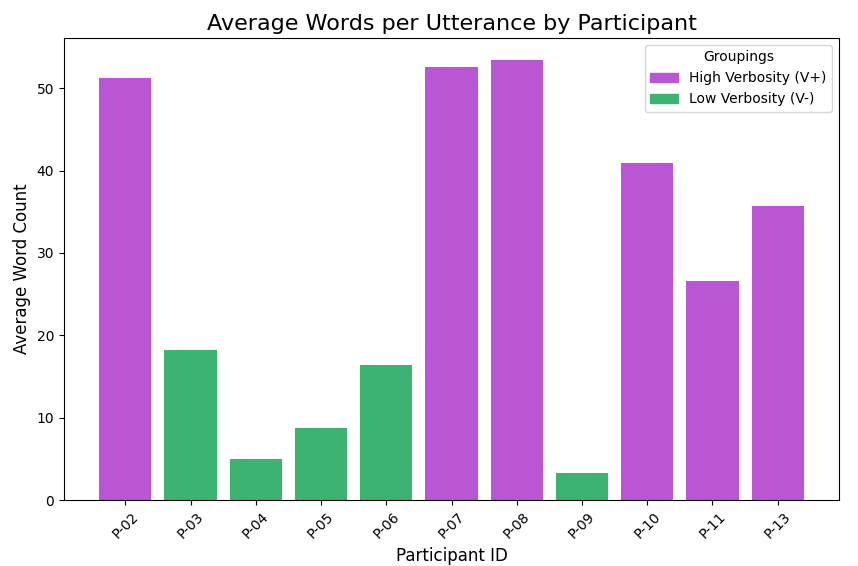}
      \caption{Average words per participant}
      \label{fig:avg_words_participant}
   \end{figure}

We noted a large difference in verbosity between the more verbose participants and the less verbose participants. Therefore, we divided the study sample into a more verbose group (\textbf{V+}) and a less verbose (\textbf{V-}) group, as distinguished in Figure \ref{fig:avg_words_participant}. Separating the study sample into \textbf{V+} and \textbf{V-} also helped to test hypothesis \textbf{H3} on potential differences in behavior between open and closed participants.

Comparing verbosity, a Mann-Whitney U Test showed participants in the \textbf{V+} group had significantly greater verbosity (average number of words spoken) than those in the \textbf{V-} group ($p < 0.01$). The \textbf{V+} group spoke an average of 43.4 words per utterance with a standard deviation of 10.9 words, while the \textbf{V-} group spoke an average of 10.3 words per utterance with a standard deviation of 6.7 words. The \textbf{V+}/\textbf{V-} split aligned with observations from video analysis of differences in noted extraversion and behavioral patterns. For example, participants in \textbf{V+} tended to give more vulnerable and open responses compared to those in \textbf{V-} (as noted by video annotators).

\subsection{Engagement and Discomfort}

Annotations from behavioral analysis demonstrated a positive average engagement in every session for every participant, even for sessions with participants who were mostly rated as displaying discomfort. During video analysis, annotators labeled events where participants appeared disengaged (-1) or engaged (1). Any portions of the video not marked as either were automatically classified as neutral behavior, with an engagement score of (0). 

\begin{figure}[thpb]
      \centering
      \includegraphics[scale=0.35]{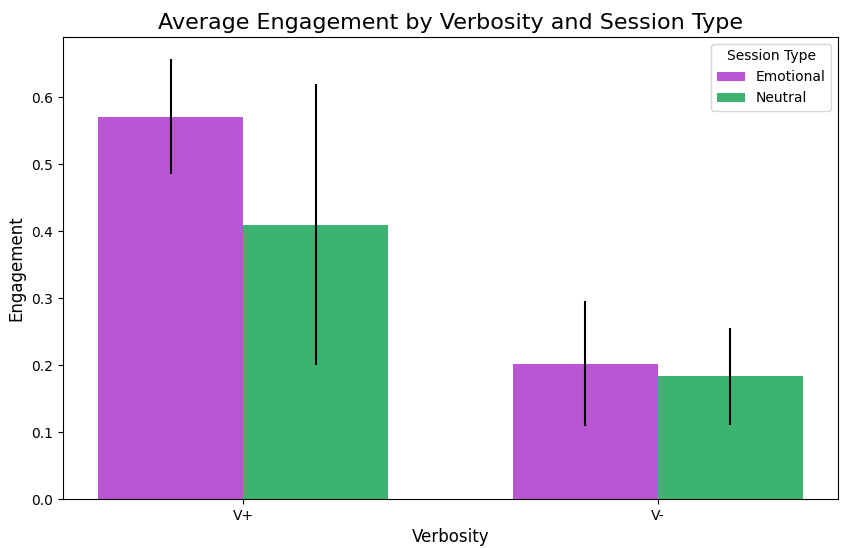}
      \caption{Engagement per verbosity group and session type}
      \label{fig:engagement}
   \end{figure}

On this scale, the \textbf{V+} group had an average engagement score of 0.57 $\pm$ 0.09 over emotional sessions and 0.41 $\pm$ 0.21 over neutral sessions, while the average engagement score of the \textbf{V-} group was 0.20 $\pm$ 0.09 over emotional sessions and 0.18 $\pm$ 0.07 over neutral sessions, as shown in Figure \ref{fig:engagement}.

Though neither group saw a \textit{significant} difference between engagement in emotional versus neutral sessions, we noted that both groups were on average more engaged in the emotional session, suggesting support for \textbf{H2}. Both groups were also on average positively engaged in all sessions.

A Mann-Whitney U Test was performed to examine differences in engagement by verbosity grouping. The \textbf{V+} group's engagement score is significantly higher ($p < 0.01$) than the the \textbf{V-} group score for average engagement in \textit{emotional} sessions. For neutral sessions, there was not a significant difference in engagement scores between groups ($p > 0.05$).

We also considered that the novelty effect \cite{shove2000defrosting, rogers2014diffusion} could artificially inflate ratings for engagement, as participants who had never seen Jibo may have been excited to interact with a new stimulus and be more engaged than they would be with a familiar stimulus. However, more than half of the participants were already familiar with Jibo due to the nature of recruiting---participants familiar with Jibo were P-02, P-05, P-08, P-09, P-10, and P-11. Four of these participants were in the \textbf{V+} group, and two were in the \textbf{V-} group.


The average engagement score of participants who had previous experience with Jibo was 0.46 $\pm$ 0.22 in the emotional sessions and 0.40 $\pm$ 0.18 in the neutral sessions, while the average engagement score of participants who were unfamiliar with Jibo was 0.34 $\pm$ 0.19 in the emotional sessions and 0.19 $\pm$ 0.16 in the neutral sessions. Participants who had never interacted with Jibo before were actually less engaged, on average, than participants who had previous familiarity with Jibo. Therefore, the novelty effect did not seem to have an effect on artificially increasing engagement.

Through qualitative analysis of participant responses and behavioral analysis, we observed that \textbf{participant engagement remained high} even during moments \textbf{when participants experienced discomfort} while sharing their feelings. 

An example of a story that caused discomfort in a participant is from P-07, who recalled: \textit{``... I felt so bad, I almost started crying. Just like this porcupine. But I didn't cry because men don't cry. And I didn't want to do it. I attract too much} [sic] \textit{people looking at me and staring and calling out so I just had to live with it.''} When sharing this emotional memory, P-07 exhibited behavior expressing discomfort (marked by annotators), like fidgeting and reduced eye contact. Despite their discomfort, they were sharing vulnerable feelings and engaging deeply with the interaction.

Another example was seen when examining utterances by P-02. They shared: \textit{``Okay, so the first day of school was a train wreck... I was like, how am I supposed to keep this up? For a year! I can't even keep it up for like a day. Because, like, when I got home, I was very mad. I was very annoyed of what would happen that day} [sic]\textit{. And I was very sad. I was eating ice cream in my blankets on my bed.''} When sharing this anecdote, P-02 was noted by video annotators to exhibit behaviors suggesting discomfort such as nervous smiling, tense shoulders, and fidgeting. However, P-02 was also clearly engaging with the activity by sharing emotions and memories that were honest and vulnerable.

Additionally, in the post-activity interview, participants acknowledged that parts of the activity made them uncomfortable, with P-10 remarking \textit{``I felt a little uncomfortable talking with a robot but I also felt really excited to talk with Jibo,''} and P-09 sharing \textit{``I was feeling, like, a little shy.''} 

\subsection{Social Robot Mitigates Discomfort}

Video annotators noted across both session types that when participants gave a response that was particularly vulnerable, honest, or open (as noted by annotators in the Openness measurement), they would often exhibit discomfort. This discomfort would continue until Jibo responded to them with either an affirmative reassurance or compliment, whereupon participants' discomfort would be reduced.

For example, after sharing a particularly vulnerable memory, P-07 recalled that they were \textit{``...Feeling like I don't really belong here, and why the hell am I even doing this?''} and immediately began exhibiting uncomfortable behaviors like flitting their eyes around the area and intense fidgeting. Once Jibo reassured P-07 by saying \textit{``I can see how this piece of art triggered some unique memories for you!''}, P-07 settled down and began making eye contact with Jibo again.

This pattern occurred in both the \textbf{V+} and \textbf{V-} groups, but the \textbf{V+} group shared more vulnerable, honest, or open answers than the \textbf{V-} group. In total, the \textbf{V+} group gave 97 total vulnerable, honest, or open answers (60 in emotional sessions and 37 in neutral sessions) and the \textbf{V-} group only gave 14 (10 in emotional sessions and 4 in neutral sessions). 

A Mann-Whitney U Test was performed to test for significant differences between the number of vulnerable answers shared in the \textbf{V+} group versus the \textbf{V-} group. The \textbf{V+} group shared significantly more vulnerable answers than the \textbf{V-} group ($p < 0.01$) in both sessions. This finding supports \textbf{H3} for emotional art, as the \textbf{V+} group was shown to both share significantly more vulnerable feelings and be significantly more engaged than the \textbf{V-} group in emotional sessions.

Of 111 total vulnerable answers across all groups, 94 answers (or 84.7\% of answers) were followed by a Jibo response that reduced the participant's discomfort noticeably (as noted by the video annotators), while 15.3\% of vulnerable answers followed by a Jibo response led to discomfort that either remained the same or increased. 

When looking at the 97 vulnerable answers given by the \textbf{V+} group, 83 (85.6\%) were followed by a Jibo utterance that visibly reduced the participant's discomfort. Though the \textbf{V+} group provided more vulnerable answers, the proportion of times their discomfort was reduced by Jibo (after sharing vulnerably) was higher than that of the entire population. 

Additionally, with 97 vulnerable answers across 6 members, the \textbf{V+} group produced an average of 16.17 vulnerable answers per participant. Each participant was only asked 24 questions in total, leading to \textbf{V+} group members sharing vulnerably in over two-thirds of their interactions with Jibo.

Examining the \textbf{V+} group further, in every session, participants' discomfort post-vulnerable response was reduced more often than not after Jibo's next utterance. There was no significant difference in the proportion of vulnerable answers that led to reduced discomfort between emotional and neutral sessions, though participants in general tended to share more vulnerable, honest, or open answers during the emotional sessions (70 answers total, 60 from \textbf{V+}) compared to the neutral sessions (41 answers total, 37 from \textbf{V+}).

Qualitative analysis of the post-interaction interview showed that participants enjoyed Jibo's responses and particularly appreciated that Jibo appeared to actively listen to them. P-07 remarked \textit{``I liked that when I say something, he really takes the time to think and he gives something corresponding to what I said''}, P-02 shared that \textit{``It was very fun to talk with something that was not human but also could probably hear me''}, and P-06 stated \textit{``I liked how Jibo responded to me and I think it understanded} [sic] \textit{me.''}

Fisher’s Exact test was used to compare instances where Jibo's responses did or did not reduce participant discomfort after vulnerable utterances. Jibo’s reassuring utterances significantly improved ($p<<.001$) participant comfort after they were vulnerable, honest, and open.


\section{DISCUSSION}
Art is a well-explored vehicle for helping children learn and practice SEL competencies \cite{elias1997promoting, eisner2002arts, brouillette2009arts}, though it has been minimally explored in tandem with social robots. Cooney et al. have examined social robots for art therapy \cite{cooney2018design, cooney2021robot}, and curricula have been designed to promote SEL skills in children when observing and reflecting on art \cite{ebert2015teaching, met_curriculum}. This study sought to expand on these prior works through a demonstration of scaffolding emotional conversations about art by using social robots for children to interactively develop and practice SEL competencies. 

\textbf{Conversations about emotional art scaffolded by social robots can foster empathy in children.} Significantly higher rates of empathetic reasoning were exhibited in emotional art versus neutral art sessions, strongly supporting \textbf{H1}. Observing and reflecting on emotional art with Jibo promoted emotional transfer and empathetic thinking, which help build empathy as an SEL competency \cite{elias1997promoting, skoog2020evidence}. The presence of Jibo and the interactive conversation facilitated by Jibo appeared to promote empathetic connection with the participant and supports the use of social robots to scaffold activities for building SEL skills. Scaffolding activities to reflect on art allowed for richer reflection on emotions observed in the art and how those emotions might connect to the participant's life. These results expand the evidence base on using social robots for teaching emotion recognition and empathy, not just for neurodivergent children \cite{kewalramani2023scoping,marino2020outcomes,yun2017social,wolfe2018deploying}, but also to provide skill-building for neurotypical children.

\textbf{Children are highly engaged in social robot-driven SEL practice, even when sharing vulnerable reflections and potentially experiencing discomfort.} On average, every participant was rated as having positive engagement (raters noted strong eye contact and deep, thoughtful contributions to the interaction) across sessions. This result demonstrated that interactions with Jibo successfully held children's interest and is consistent with previous findings on how social robots can promote user engagement \cite{westlund2017children, fridin2014storytelling}. Furthermore, participants who had previously interacted with Jibo had higher average engagement scores than those without prior experience, suggesting that engagement was not due to the novelty effect \cite{shove2000defrosting, rogers2014diffusion}. This higher engagement from participants with past experience may suggest a self-selection bias in study participation but also points to the potential for leveraging social robots for longitudinal SEL programming, where continuous interaction may lead to higher engagement. From qualitative analysis of utterances, participant engagement remained high even during moments of discomfort, suggesting that participants felt it was a safe space to feel the discomfort that arises from vulnerability and could continue to engage with the robot. Results support \textbf{H2} for the \textbf{V+} group, as participants' average engagement levels in the emotional sessions were higher than in neutral sessions. However, the difference was not statistically significant, and more data is needed to reach a conclusive result. Results also support \textbf{H3} for emotional art, as the \textbf{V+} group, who were significantly more open, had significantly higher levels of engagement than the \textbf{V-} group in the emotional session.

\textbf{A social robot can help mitigate the discomfort a child feels when sharing vulnerable feelings.} Discomfort that arose during and after participants shared vulnerable feelings decreased significantly after the robot offered a reassuring  response. These findings demonstrated that interacting with Jibo was a comforting experience, consistent with previous findings showing that social robots can reduce children's anxiety and promote comfort and disclosure \cite{dosso2023safe}. Additionally, \textbf{V+} group members on average shared deeply and openly 16.17 times out of 24 utterances total (approximately two-thirds of utterances). Participants appeared to share deeply as they felt comfortable around Jibo due to how his responses were personalized to their utterances, which may have helped the participants feel listened to and cared for. One change that could help the \textbf{V-} group share more is using the robot to detect when further questioning is helpful---for example, when a participant said ``I don't know'', Jibo would move on and ask the participant what was confusing. The participant may have been able to share more if Jibo had instead prompted them to think again about their feelings.

This study was limited by a small sample size, and future works will expand to a larger, more diverse population to validate findings. Behavioral analysis also suggested that a laboratory setting may have inadvertently heightened discomfort, as participants disclosed vulnerable information amidst strangers in an unfamiliar place. Conducting future studies in familiar areas could encourage more open and vulnerable responses. Future research would also benefit from exploring longitudinal interactions to better understand changes over time in participants' SEL skills. This study only investigated reactions to three pieces of art representing three emotions. However, future research should include a broader spectrum of art and a wider range of emotions to facilitate more comprehensive exploration of emotional responses. 

\section{CONCLUSION}
We explored how social robots can foster social-emotional learning (SEL) competencies in children through conversations about art. Our investigation involved 11 participants who engaged in two sessions discussing emotional and neutral artworks, facilitated by social robot scaffolding.

Findings demonstrated that discussing emotional art with a social robot is an effective method for emotional self-awareness and empathy (key SEL skills). Reflecting on art prompted children to engage deeply and thoughtfully with the social robot, and it was able to alleviate discomfort to encourage continued engagement and emotional exploration.


This work demonstrated a promising method of fostering social-emotional learning in children and provides an initial foundation for future inclusive, expansive, and longitudinal studies to validate and expand the capabilities for robot- and art-mediated interactions for building SEL skills. 








\bibliographystyle{IEEEtran}
\bibliography{IEEEabrv,SEL-Jibo}

\begin{thebibliography}{10}
\providecommand{\url}[1]{#1}
\csname url@rmstyle\endcsname
\providecommand{\newblock}{\relax}
\providecommand{\bibinfo}[2]{#2}
\providecommand\BIBentrySTDinterwordspacing{\spaceskip=0pt\relax}
\providecommand\BIBentryALTinterwordstretchfactor{4}
\providecommand\BIBentryALTinterwordspacing{\spaceskip=\fontdimen2\font plus
\BIBentryALTinterwordstretchfactor\fontdimen3\font minus \fontdimen4\font\relax}
\providecommand\BIBforeignlanguage[2]{{%
\expandafter\ifx\csname l@#1\endcsname\relax
\typeout{** WARNING: IEEEtran.bst: No hyphenation pattern has been}%
\typeout{** loaded for the language `#1'. Using the pattern for}%
\typeout{** the default language instead.}%
\else
\language=\csname l@#1\endcsname
\fi
#2}}

\bibitem{elias1997promoting}
M.~J. Elias, M.~Elias, J.~E. Zins, and R.~P. Weissberg, \emph{Promoting social and emotional learning: Guidelines for educators}.\hskip 1em plus 0.5em minus 0.4em\relax Ascd, 1997.

\bibitem{skoog2020evidence}
A.~Skoog-Hoffman, C.~Ackerman, A.~Boyle, H.~Schwartz, B.~Williams, R.~Jagers, L.~Dusenbury, M.~Greenberg, J.~Mahoney, K.~Schonert-Reichl, \emph{et~al.}, ``Evidence-based social and emotional learning programs: Casel criteria updates and rationale,'' \emph{Retrieved February}, vol.~6, p. 2023, 2020.

\bibitem{greenberg2017social}
M.~T. Greenberg, C.~E. Domitrovich, R.~P. Weissberg, and J.~A. Durlak, ``Social and emotional learning as a public health approach to education,'' \emph{The future of children}, pp. 13--32, 2017.

\bibitem{mondi_fostering_2021}
C.~F. Mondi, A.~Giovanelli, and A.~J. Reynolds, ``Fostering socio-emotional learning through early childhood intervention,'' \emph{International Journal of Child Care and Education Policy}, vol.~15, no.~1, pp. 1--43, 2021.

\bibitem{hawkins2008effects}
J.~D. Hawkins, R.~Kosterman, R.~F. Catalano, K.~G. Hill, and R.~D. Abbott, ``Effects of social development intervention in childhood 15 years later,'' \emph{Archives of pediatrics \& adolescent medicine}, vol. 162, no.~12, pp. 1133--1141, 2008.

\bibitem{taylor_promoting_2017}
R.~D. Taylor, E.~Oberle, J.~A. Durlak, and R.~P. Weissberg, ``\BIBforeignlanguage{en}{Promoting {Positive} {Youth} {Development} {Through} {School}-{Based} {Social} and {Emotional} {Learning} {Interventions}: {A} {Meta}-{Analysis} of {Follow}-{Up} {Effects}},'' \emph{\BIBforeignlanguage{en}{Child Development}}, vol.~88, no.~4, pp. 1156--1171, 2017.

\bibitem{durlak_impact_2011}
J.~A. Durlak, R.~P. Weissberg, A.~B. Dymnicki, R.~D. Taylor, and K.~B. Schellinger, ``\BIBforeignlanguage{en}{The {Impact} of {Enhancing} {Students}’ {Social} and {Emotional} {Learning}: {A} {Meta}-{Analysis} of {School}-{Based} {Universal} {Interventions}},'' \emph{\BIBforeignlanguage{en}{Child Development}}, vol.~82, no.~1, pp. 405--432, 2011.

\bibitem{bridgeland2013missing}
J.~Bridgeland, M.~Bruce, and A.~Hariharan, ``The missing piece: A national teacher survey on how social and emotional learning can empower children and transform schools. a report for casel.'' \emph{Civic Enterprises}, 2013.

\bibitem{aspen2019nation}
{Aspen Institute National Commission on Social, Emotional, and Academic Development}, ``From a nation at risk to a nation at hope: Recommendations from the national commission on social, emotional, \& academic development.''\hskip 1em plus 0.5em minus 0.4em\relax Aspen Institute Washington, DC, 2019.

\bibitem{kanda2012children}
T.~Kanda, M.~Shimada, and S.~Koizumi, ``Children learning with a social robot,'' in \emph{Proceedings of the 7th annual ACM/IEEE international conference on Human-Robot Interaction}, 2012, pp. 351--358.

\bibitem{thomaz2013active}
A.~L. Thomaz, M.~Cakmak, and K.~Clark, ``Active social learning in humans and robots,'' \emph{Social learning theory: Phylogenetic considerations across animal, plant, and microbial taxa}, pp. 113--28, 2013.

\bibitem{kanero2018social}
J.~Kanero, V.~Ge{\c{c}}kin, C.~Oran{\c{c}}, E.~Mamus, A.~C. K{\"u}ntay, and T.~G{\"o}ksun, ``Social robots for early language learning: Current evidence and future directions,'' \emph{Child Development Perspectives}, vol.~12, no.~3, pp. 146--151, 2018.

\bibitem{van2019social}
R.~Van~den Berghe, J.~Verhagen, O.~Oudgenoeg-Paz, S.~Van~der Ven, and P.~Leseman, ``Social robots for language learning: A review,'' \emph{Review of Educational Research}, vol.~89, no.~2, pp. 259--295, 2019.

\bibitem{neumann2020social}
M.~M. Neumann, ``Social robots and young children’s early language and literacy learning,'' \emph{Early Childhood Education Journal}, vol.~48, no.~2, pp. 157--170, 2020.

\bibitem{chen2020teaching}
H.~Chen, H.~W. Park, and C.~Breazeal, ``Teaching and learning with children: Impact of reciprocal peer learning with a social robot on children’s learning and emotive engagement,'' \emph{Computers \& Education}, vol. 150, p. 103836, 2020.

\bibitem{williams2018popbots}
R.~Williams, ``Popbots: leveraging social robots to aid preschool children's artificial intelligence education,'' Ph.D. dissertation, Massachusetts Institute of Technology, 2018.

\bibitem{rafique2020computation}
M.~Rafique, M.~A. Hassan, A.~Jaleel, H.~Khalid, and G.~Bano, ``A computation model for learning programming and emotional intelligence,'' \emph{IEEE Access}, vol.~8, pp. 149\,616--149\,629, 2020.

\bibitem{westlund2017children}
J.~M.~K. Westlund, L.~Dickens, S.~Jeong, P.~L. Harris, D.~DeSteno, and C.~L. Breazeal, ``Children use non-verbal cues to learn new words from robots as well as people,'' \emph{International Journal of Child-Computer Interaction}, vol.~13, pp. 1--9, 2017.

\bibitem{fridin2014storytelling}
M.~Fridin, ``Storytelling by a kindergarten social assistive robot: A tool for constructive learning in preschool education,'' \emph{Computers \& education}, vol.~70, pp. 53--64, 2014.

\bibitem{dosso2023safe}
J.~A. Dosso, J.~N. Kailley, S.~E. Martin, and J.~M. Robillard, ``“a safe space for sharing feelings”: perspectives of children with lived experiences of anxiety on social robots,'' \emph{Multimodal Technologies and Interaction}, vol.~7, no.~12, p. 118, 2023.

\bibitem{pashevich2022can}
E.~Pashevich, ``Can communication with social robots influence how children develop empathy? best-evidence synthesis,'' \emph{AI \& SOCIETY}, vol.~37, no.~2, pp. 579--589, 2022.

\bibitem{spitale2022socially}
M.~Spitale, S.~Okamoto, M.~Gupta, H.~Xi, and M.~J. Matari{\'c}, ``Socially assistive robots as storytellers that elicit empathy,'' \emph{ACM Transactions on Human-Robot Interaction (THRI)}, vol.~11, no.~4, pp. 1--29, 2022.

\bibitem{hurst2020social}
N.~Hurst, C.~Clabaugh, R.~Baynes, J.~Cohn, D.~Mitroff, and S.~Scherer, ``Social and emotional skills training with embodied moxie,'' \emph{arXiv preprint arXiv:2004.12962}, 2020.

\bibitem{kewalramani2023scoping}
S.~Kewalramani, K.-A. Allen, E.~Leif, and A.~Ng, ``A scoping review of the use of robotics technologies for supporting social-emotional learning in children with autism,'' \emph{Journal of Autism and Developmental Disorders}, pp. 1--15, 2023.

\bibitem{marino2020outcomes}
F.~Marino, P.~Chil{\`a}, S.~T. Sfrazzetto, C.~Carrozza, I.~Crimi, C.~Failla, M.~Bus{\`a}, G.~Bernava, G.~Tartarisco, D.~Vagni, \emph{et~al.}, ``Outcomes of a robot-assisted social-emotional understanding intervention for young children with autism spectrum disorders,'' \emph{Journal of autism and developmental disorders}, vol.~50, pp. 1973--1987, 2020.

\bibitem{yun2017social}
S.-S. Yun, J.~Choi, S.-K. Park, G.-Y. Bong, and H.~Yoo, ``Social skills training for children with autism spectrum disorder using a robotic behavioral intervention system,'' \emph{Autism Research}, vol.~10, no.~7, pp. 1306--1323, 2017.

\bibitem{wolfe2018deploying}
E.~Wolfe, J.~Weinberg, and S.~Hupp, ``Deploying a social robot to co-teach social emotional learning in the early childhood classroom,'' in \emph{Proceedings of the 13th Annual ACM/IEEE International Conference on Human--Robot Interaction, Chicago, IL, USA}, 2018, pp. 5--8.

\bibitem{eisner2002arts}
E.~W. Eisner, \emph{The arts and the creation of mind}.\hskip 1em plus 0.5em minus 0.4em\relax Yale University Press, 2002.

\bibitem{brouillette2009arts}
L.~Brouillette, ``How the arts help children to create healthy social scripts: Exploring the perceptions of elementary teachers,'' \emph{Arts Education Policy Review}, vol. 111, no.~1, pp. 16--24, 2009.

\bibitem{farrington2019arts}
C.~A. Farrington, J.~Maurer, M.~R.~A. McBride, J.~Nagaoka, J.~Puller, S.~Shewfelt, E.~M. Weiss, and L.~Wright, ``Arts education and social-emotional learning outcomes among k-12 students: Developing a theory of action.'' \emph{University of Chicago Consortium on School Research}, 2019.

\bibitem{cooney2018design}
M.~D. Cooney and M.~L.~R. Menezes, ``Design for an art therapy robot: An explorative review of the theoretical foundations for engaging in emotional and creative painting with a robot,'' \emph{Multimodal Technologies and Interaction}, vol.~2, no.~3, p.~52, 2018.

\bibitem{cooney2021robot}
M.~Cooney, ``Robot art, in the eye of the beholder?: Personalized metaphors facilitate communication of emotions and creativity,'' \emph{Frontiers in Robotics and AI}, vol.~8, p. 668986, 2021.

\bibitem{heath1998living}
S.~B. Heath, E.~Soep, and A.~Roach, \emph{Living the arts through language+ learning: A report on community-based youth organizations}.\hskip 1em plus 0.5em minus 0.4em\relax Americans for the Arts, 1998.

\bibitem{eddy2021local}
M.~Eddy, C.~Blatt-Gross, S.~N. Edgar, A.~Gohr, E.~Halverson, K.~Humphreys, and L.~Smolin, ``Local-level implementation of social emotional learning in arts education: Moving the heart through the arts,'' \emph{Arts Education Policy Review}, vol. 122, no.~3, pp. 193--204, 2021.

\bibitem{ebert2015teaching}
M.~Ebert, J.~D. Hoffmann, Z.~Ivcevic, C.~Phan, and M.~A. Brackett, ``Teaching emotion and creativity skills through art: A workshop for children,'' \emph{The International Journal of Creativity \& Problem Solving}, vol.~25, no.~2, pp. 23--35, 2015.

\bibitem{met_curriculum}
``Social and emotional learning through art,'' Online Curriculum, The Metropolitan Museum of Art, New York, New York, NY, USA, 2022.

\bibitem{jibo}
J.~Inc., ``{Jibo},'' \url{https://jibo.com}, [Online; accessed 01-August-2023].

\bibitem{cordero2022review}
J.~R. Cordero, T.~R. Groechel, and M.~J. Matari{\'c}, ``A review and recommendations on reporting recruitment and compensation information in hri research papers,'' in \emph{2022 31st IEEE International Conference on Robot and Human Interactive Communication (RO-MAN)}.\hskip 1em plus 0.5em minus 0.4em\relax IEEE, 2022, pp. 1627--1633.

\bibitem{mchugh_interrater_2012}
M.~L. McHugh, ``Interrater reliability: the kappa statistic,'' \emph{Biochemia medica}, vol.~22, no.~3, pp. 276--282, 2012.

\bibitem{shove2000defrosting}
E.~Shove and D.~Southerton, ``Defrosting the freezer: From novelty to convenience: A narrative of normalization,'' \emph{Journal of Material Culture}, vol.~5, no.~3, pp. 301--319, 2000.

\bibitem{rogers2014diffusion}
E.~M. Rogers, A.~Singhal, and M.~M. Quinlan, ``Diffusion of innovations,'' in \emph{An integrated approach to communication theory and research}.\hskip 1em plus 0.5em minus 0.4em\relax Routledge, 2014, pp. 432--448.

\end{thebibliography}

\end{document}